\documentclass[11pt]{article}

\usepackage[preprint]{acl}
\usepackage{times}
\usepackage{latexsym}

\usepackage[T1]{fontenc}
\usepackage[utf8]{inputenc}

\usepackage{microtype}

\usepackage{inconsolata}

\usepackage{graphicx}
\usepackage{multirow}
\usepackage{booktabs}
\usepackage{adjustbox}
\usepackage{subcaption}
\usepackage{geometry}
\usepackage{hyperref}
\usepackage{fontawesome}

\title{\textit{%
\raisebox{-0.45ex}{\includegraphics[height=2\fontcharht\font`\B]{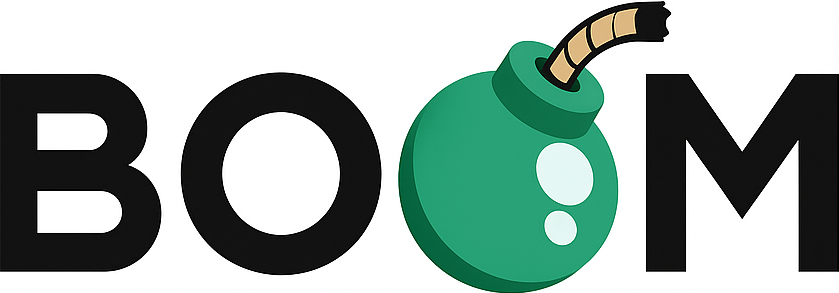}}%
~: Beyond Only One Modality}\\
KIT's Multimodal Multilingual Lecture Companion}

\author{\textbf{Sai Koneru},
  \textbf{Fabian Retkowski},
  \textbf{Christian Huber},
  \textbf{Lukas Hilgert},\\
  \textbf{Seymanur Akti},
  \textbf{Enes Yavuz Ugan},
  \textbf{Alexander Waibel}, 
  \textbf{Jan Niehues}
 \\
 Karlsruhe Institute of Technology \\
 \\
   \href{mailto:email@domain}{firstname.lastname@kit.edu}
  }

\begin{document}
\maketitle
\begin{abstract}
The globalization of education and rapid growth of online learning have made localizing educational content a critical challenge. Lecture materials are inherently multimodal, combining spoken audio with visual slides, which requires systems capable of processing multiple input modalities. To provide an accessible and complete learning experience, translations must preserve all modalities: text for reading, slides for visual understanding, and speech for auditory learning. We present \textbf{BOOM}, a multimodal multilingual lecture companion that jointly translates lecture audio and slides to produce synchronized outputs across three modalities: translated text, localized slides with preserved visual elements, and synthesized speech. This end-to-end approach enables students to access lectures in their native language while aiming to preserve the original content in its entirety. Our experiments demonstrate that slide-aware transcripts also yield cascading benefits for downstream tasks such as summarization and question answering. The demo video and code can be found at \url{https://ai4lt.github.io/boom/}\footnote{All released code and models are licensed under the MIT License}.
\end{abstract}

\section{Introduction}

Access to educational content in a learner’s native language greatly enhances the learning experience for university students. Localizing lecture material reduces communication barriers, improves accessibility, and enables learners to engage more deeply with complex concepts. As higher education becomes increasingly global, the ability to provide multilingual lecture content both in-person and online has become essential to increase accessibility to educational resources \citep{muthuswamy2023analysing, gambier2023audiovisual}.

With the ongoing digitalization of teaching, lecture content itself is inherently multimodal. The primary modality is the lecture audio, which can be converted into transcripts via Automatic Speech Recognition (ASR) \cite{pham2019very, radford2022whisper}. Instructional material is presented through slides, and additional outputs, such as summaries, chapters, and question–answer interactions, can be generated based on the transcript in a cascaded setup using modern Large Language Model (LLM)-based systems to enhance the learning experience \cite{waibel2012simultaneous, waibel2014translation, anderer_2025,retkowski-etal-2025-summarizing}. To ensure accessibility for all students, including non-native speakers, these outputs should also be available in multiple languages. Effective localization must therefore handle this diversity of content, spanning audio, text, and visual materials, making lecture translation a truly multimodal challenge.

\begin{figure*}[ht]
    \centering
    \begin{subfigure}{\columnwidth}
        \centering
        \includegraphics[width=\linewidth,trim={7.9cm 4.8cm 7.95cm 3.78cm},clip]{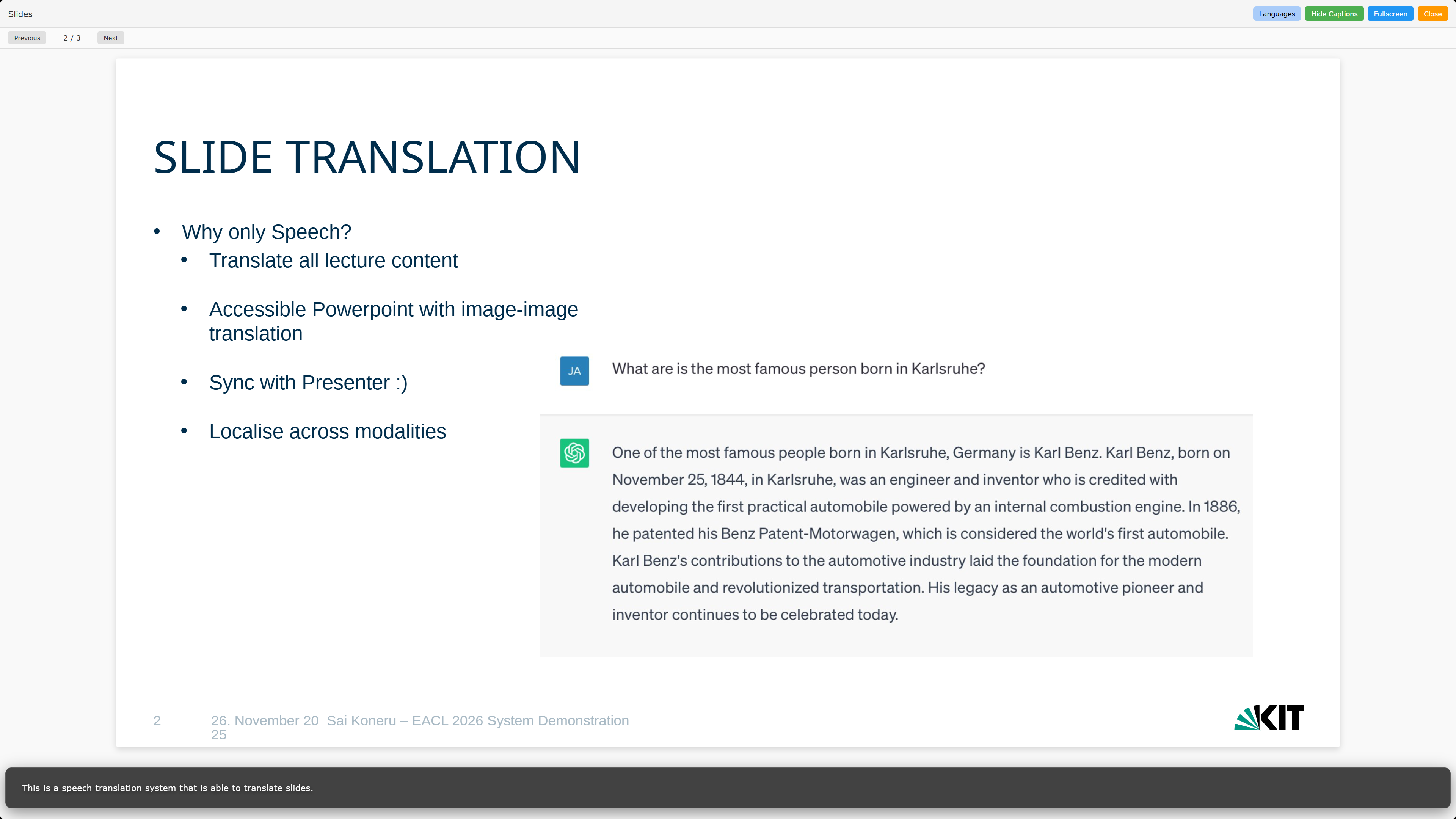}
        \caption{Original English Slide}
        \label{fig:slide_en}
    \end{subfigure}
    \hfill
    \begin{subfigure}{\columnwidth}
        \centering
        \includegraphics[width=\linewidth,trim={7.9cm 4.8cm 7.95cm 3.78cm},clip]{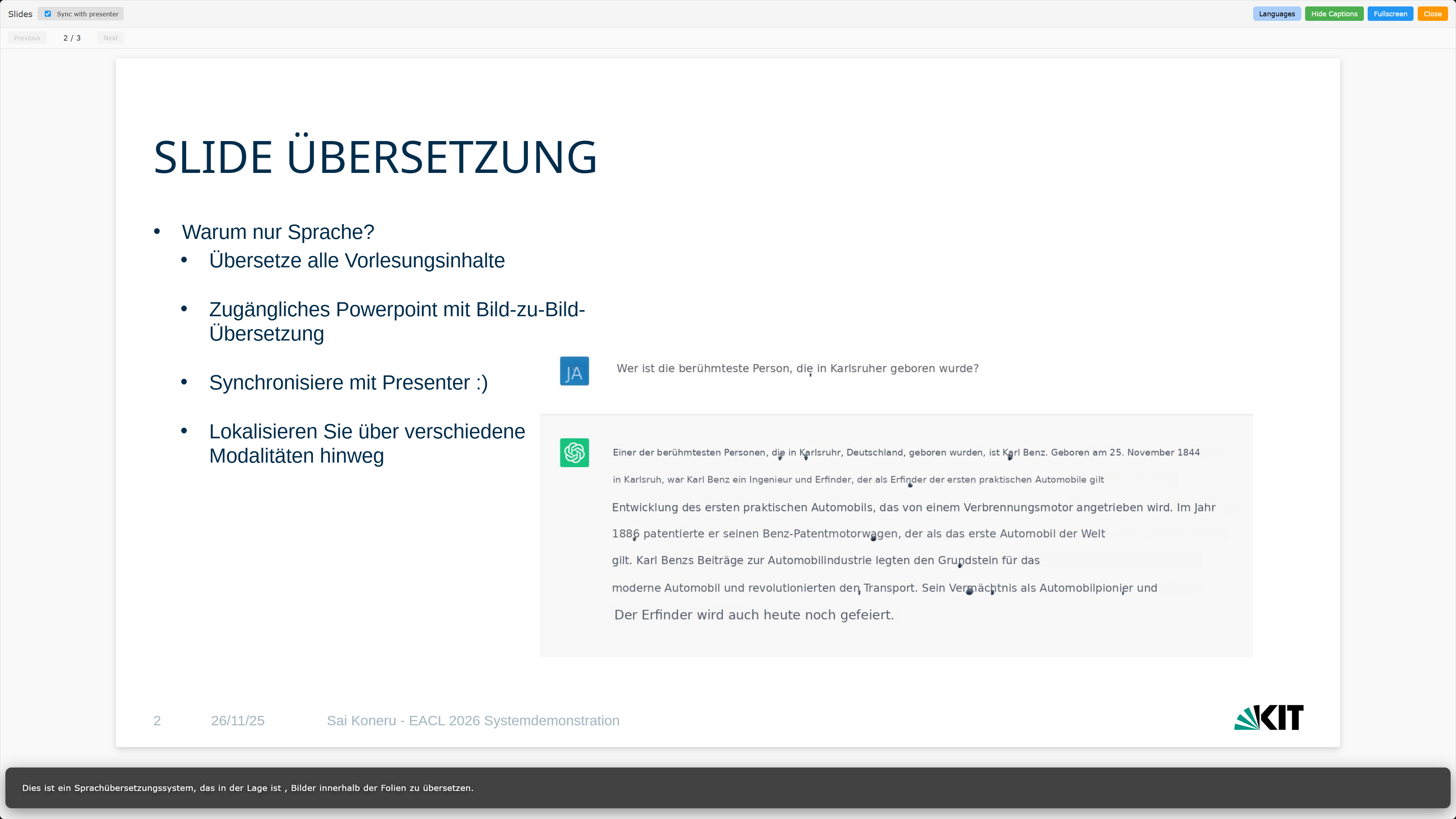}
        \caption{Translated German Slide}
        \label{fig:slide_de}
    \end{subfigure}
    \caption{Comparison of the English (original) and German (translated) slides. Text outside the images is translated with a unimodal system for efficiency, while text inside the images is translated using a multimodal system.}
    \label{fig:slides_side_by_side}
\end{figure*}

This multimodality introduces complexity but also offers valuable contextual signals. Images often contain additional cues ranging from scene information in natural images and definitions, formulas, diagrams, and domain-specific terminology in slides that help disambiguate spoken content \citep{nguyen2025cocktail} and support downstream tasks such as Summarization (SUM) and Question Answering (QA). Leveraging these visual cues enables translation systems to move beyond audio-only processing and incorporate richer semantic information throughout the lecture translation pipeline \citep{waibel2018translation,chen-etal-2024-m3av,Sinhamahapatra2025-cf}.

Machine Translation (MT) forms the foundation of localization, evolving from rule-based systems \citep{hutchins2004georgetown} to Neural MT (NMT; \citealt{vaswani2017attention, koehn2017six, johnson2017google}) and then Speech Translation (ST), which directly translates spoken content. Modern ST handles many languages \citep{barrault2023seamless} but often processes short segments, limiting context and potential to benefit from multimodality. 

In this work, we address multimodality on both the input and output sides of lecture localization. On the input side, we incorporate slide screenshots into the ST pipeline to provide contextual grounding that improves translation accuracy and downstream LLM performance. On the output side, we tackle the challenge of localizing lecture slides themselves. Slides often contain text embedded within images, such as diagram labels, equations, or annotations, that existing ST tools typically ignore. Localizing such material requires detecting, recognizing, translating, and re-rendering text while preserving layout, alignment, font style, and visual coherence (illustrated in Figure \ref{fig:slides_side_by_side}).

To overcome these limitations, we extend the Lecture Translator (LT) software \citep{huber2023end} with OmniFusion \citep{koneru2025omnifusionsimultaneousmultilingualmultimodal}, a multilingual multimodal ST model that uses slide images to enrich translation. We further introduce a fully open-source slide translation system capable of translating text inside slide images and rendering it back into its original layout, enabling complete slide localization. Together, these components form a unified multimodal lecture localization pipeline that combines improved ST with synchronized slide translation, significantly enhancing accessibility for learners across languages.

Our main contributions include:

\begin{itemize}

\item Adapt and integrate OmniFusion to leverage lecture slide screenshots during live translation, by extracting relevant slides from segmented audio.

\item Introduce an open-source image-to-image translation pipeline with modular components, enabling future research on full-image/slide translation and rendering.

\item Demonstrate the impact of including images on ST for downstream NLP tasks across different LLMs, showing performance improvements in different language pairs. We also evaluate several optical character recognition (OCR) models and the translation quality of unimodal and multimodal NMT models for image translation.

\end{itemize}

\begin{figure*}[!ht]
    \centering
    \includegraphics[width=1\linewidth]{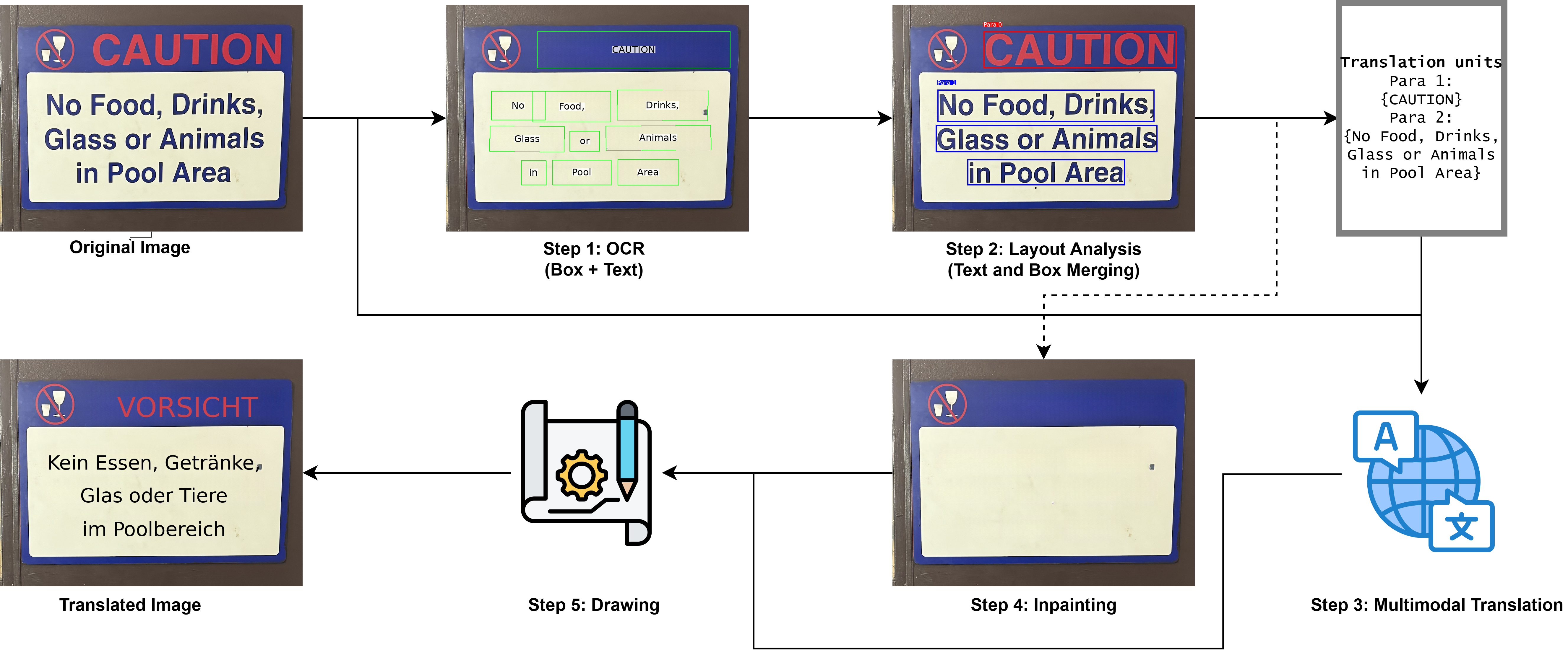}
    \caption{Overview of the image translator pipeline. Arrows indicate the inputs to each step. All steps are model-based except for drawing, which uses heuristic rules.}
    \label{fig:pipeline}
\end{figure*}

\section{System Description}

To fully localize lecture content, including audio and slides, across multiple modalities and languages, and to support accessibility tasks such as SUM and QA, we develop multimodal translation systems. Our approach performs multimodal ST and leverages the resulting transcripts for downstream LLM tasks. We also translate slides by converting text and images into the target language while preserving their layout and visual coherence.

To map visual context to each audio segment and improve usability, we built a PDF viewer that displays slides with overlaid captions synchronized to the presenter’s selected slide (Figure \ref{fig:clientview_de}). This interface enables participants to follow translations while viewing slides and allows the system to automatically identify which slide corresponds to each audio segment, providing essential context for multimodal ST.

In this section, we first describe the multimodal ST pipeline, including how slide images are extracted and associated with audio segments. Next, we outline how the resulting translations are used for downstream SUM and QA. Finally, we present the slide-translation process, detailing how text embedded within slide images is detected, translated, and re-rendered. Additionally, details about Text-to-Speech (TTS) are described in Appendix \ref{sec:tts}.

\subsection{Multimodal Speech Translation}

Several ST systems support translation across multiple languages, but they are not directly suitable for live lectures. Most are trained for offline tasks with fixed segmentation, which is incompatible with streaming audio, and require simultaneous translation policies \citep{niehues2016dynamic, niehues2018low, polak2023incremental} to determine when enough audio has been received. Existing systems are either unimodal, ignoring slides, or multimodal but lack multilingual support. Lecture scenarios demand both multimodality and multilinguality.

To address these challenges, we adopt the OmniFusion model for multimodal ST, which supports multiple languages and has been shown to improve quality when integrating slides. Since it is trained primarily on clean speech, we fine-tune it on noisy data\footnote{\url{https://huggingface.co/skoneru/OmniFusion_v2}}. For streaming translation, we follow the LT policy \citep{huber2023end}, combining voice-activity detection with Local-Agreement to produce low-latency outputs.

Accurate visual context is crucial for effective translation. The PDF viewer tracks the slide displayed during each audio segment, allowing us to extract a screenshot from the middle of the segment and feed it to the ST model. This provides relevant visual cues, improving translation quality, especially for technical content, while enabling participants to follow translations in real time.

\begin{table*}[!ht]
\centering
\begin{tabular}{@{}ccccccc@{}}
\toprule
Model          & CER ($\downarrow$) & TER ($\downarrow$) & Sub. & Del.  & Ins. & \begin{tabular}[c]{@{}c@{}}Average Time ($\downarrow$)\\ (Seconds)\end{tabular} \\ \midrule
EasyOCR        & 56.44              & 57.44              & 1488 & 29337 & 553  & 0.22                                                                            \\
Paddle-OCR-v4  & 11.31              & 16.53              & 880  & 2791  & 2435 & 0.06                                                                            \\
Paddle-OCR-v5  & 13.48              & 16.91              & 1717 & 2639  & 3014 & 0.10                                                                            \\
Qwen-2.5-VL 7B & 13.54              & 12.77              & 413  & 2348  & 3144 & 5.10                                                                            \\ \bottomrule
\end{tabular}
\caption{Performance of OCR models on the VISTRA benchmark. Evaluations are restricted to English text in signboards and similar visual contexts, and therefore do not reflect performance across broader OCR domains.}
\label{tab:ocr}
\end{table*}

\subsubsection{Summarization \& Question-Answering}

Beyond translating spoken content, lecture material should also be chaptered \cite{zechner2000diasumm, zechner2000minimizing,schneider2025policies,retkowski-waibel-2024-text}, meaning split into coherent functional and semantic sections, and then summarized in multiple languages and made available for interactive QA. To support these tasks, we use the transcribed multimodal ST output as context. Although modern LLMs can handle long contexts efficiently, their context window is still limited, so we adopt the following strategy.

For summarization, lectures are first translated into multiple languages. Each lecture is then divided into chapters, which prevents context-window overflow and also produces conceptually cleaner summaries, since chapters contain locally coherent content and avoid the topic drift that often appears in global summaries. For each chapter, we generate several forms of compressed representations. These include transcript compressions at multiple ratios such as 50 percent, 70 percent, and 90 percent, as well as length-controlled summaries whose size is determined by the length of the source section \cite{retkowski-waibel-2025-zero}. All summaries are first produced in English to benefit from the stronger performance of LLMs on English text and are then translated into the target languages.

For QA, we follow a similar approach: the English transcript, organized by chapters, is used with Retrieval-Augmented Generation (RAG) to query an LLM \cite{anderer_2025}, and the resulting answers are translated into the target languages.

\subsection{Slide Translation}

Another challenge for making lectures accessible is translating slides into multiple languages. Slides contain both editable text and images with embedded text. For editable text, we use a Python-based PowerPoint parser\footnote{\url{https://pypi.org/project/python-pptx/}} to extract text blocks and translate them with standard unimodal MT, avoiding multimodal models due to computational cost.

Text inside images cannot be directly extracted, often lacks surrounding linguistic context, and relies on visual elements for interpretation, making multimodal translation necessary. After translation, text must be reinserted into the original image to preserve layout and visual meaning. To address this, we propose an \textbf{image-translation pipeline} that detects, recognizes, translates, and re-renders text within slide images (Figure \ref{fig:pipeline}).

\subsubsection{Optical Character Recognition}

The system begins with extracting text from slide images using PaddleOCR v5 \citep{cui2025paddleocr30technicalreport}, which supports multiple languages and outputs both recognized text and bounding boxes, typically at the word or character level. While sufficient for translation, these detections do not form coherent segments or preserve semantic structure, requiring layout analysis.

\subsubsection{Layout Analysis}

We then apply layout analysis using the Hi-SAM model\footnote{\textit{sam\_vit\_l\_0b3195.pth}} \citep{ye2025hi}, which predicts block-level regions and their constituent lines. OCR boxes are grouped into block-level and line-level segments, producing sentence-like units suitable for translation. Layout analysis also preserves structural cues, such as grouping, font size, and color, that aid re-rendering. For instance, bullet list items or diagram labels are grouped to maintain consistent formatting.

\subsubsection{Multimodal Translation}

Text from each block is concatenated and translated using OmniFusion adapted from Qwen Omni 2.5 7B\citep{ye2025omnivinci} and SeedX PPO 7B \citep{Cheng2025-id}, which leverages the slide image as visual context. This multimodal approach is particularly helpful for short, ambiguous, or visually grounded text.

\subsubsection{Inpainting}

Before inserting the translated text, the original text regions are removed using Simple-LaMa\footnote{\url{https://github.com/enesmsahin/simple-lama-inpainting/}} \citep{suvorov2021resolution}, a lightweight inpainting model that reconstructs the background with minimal artifacts, preserving slide quality.

\subsubsection{Drawing}

Translated text is then rendered back onto the slide. Fully automatic diffusion-based methods proved unsuitable because repeated edits gradually degraded clarity. Instead, a heuristic drawing module estimates original text styling and positions the translated text within the same layout and line structure. This preserves alignment, spatial organization, and overall visual coherence, ensuring the localized slide matches the structure and intent of the original.

\begin{table*}[!ht]
\centering
\resizebox{2\columnwidth}{!}{
\begin{tabular}{@{}ccccccccccccc@{}}
\toprule
\multirow{2}{*}{Model} & \multicolumn{3}{c}{de} & \multicolumn{3}{c}{es} & \multicolumn{3}{c}{ru} & \multicolumn{3}{c}{zh} \\ 
\cmidrule(l){2-13}
& BLEU ($\uparrow$) & ChrF ($\uparrow$) & COMET ($\uparrow$)
& BLEU ($\uparrow$) & ChrF ($\uparrow$) & COMET ($\uparrow$)
& BLEU ($\uparrow$) & ChrF ($\uparrow$) & COMET ($\uparrow$)
& BLEU ($\uparrow$) & ChrF ($\uparrow$) & COMET ($\uparrow$) \\ 
\midrule

\multicolumn{13}{c}{\textit{OCR Predicted + Line-level}} \\ \midrule

SeedX 7B PPO
& \phantom{0}6.7 & 21.3 & 50.9
& 18.3 & 48.8 & 68.9
& 10.8 & 37.8\textsuperscript{\rlap{*}} & 65.6\textsuperscript{\rlap{*}}
& \phantom{0}0.6 & \phantom{0}7.4 & 62.8 \\

Tower-Instruct 7B
& \phantom{0}4.5 & 23.3 & 50.5
& 11.6 & 40.0 & 63.2
& \phantom{0}7.3 & 28.9 & 59.1
& \phantom{0}3.5\textsuperscript{\rlap{*}} & 17.2 & 63.1 \\

OmniFusion
& \phantom{0}9.2\textsuperscript{\rlap{*}} & 25.3\textsuperscript{\rlap{*}} & 53.5\textsuperscript{\rlap{*}}
& 19.8\textsuperscript{\rlap{*}} & 50.7\textsuperscript{\rlap{*}} & 70.4\textsuperscript{\rlap{*}}
& 11.0\textsuperscript{\rlap{*}} & 34.8 & 64.6
& \phantom{0}1.3 & 22.1\textsuperscript{\rlap{*}} & 67.6\textsuperscript{\rlap{*}} \\

\midrule
\multicolumn{13}{c}{\textit{OCR Predicted + Layout-level}} \\ \midrule

SeedX 7B PPO
& 10.3 & 23.7 & 53.1
& 28.4\textsuperscript{\rlap{*}} & 56.8\textsuperscript{\rlap{*}} & 74.0
& 17.3\textsuperscript{\rlap{*}} & 43.4\textsuperscript{\rlap{*}} & 71.2\textsuperscript{\rlap{*}}
& \phantom{0}2.0 & 14.8 & 67.8 \\

Tower-Instruct 7B
& 11.2 & 27.4 & 53.7
& 19.1 & 46.4 & 68.2
& 10.6 & 30.7 & 63.3
& \phantom{0}8.4\textsuperscript{\rlap{*}} & 22.9 & 68.3 \\

OmniFusion
& 13.6\textsuperscript{\rlap{*}} & 30.1\textsuperscript{\rlap{*}} & 56.9\textsuperscript{\rlap{*}}
& 28.1 & 56.2 & 74.5\textsuperscript{\rlap{*}}
& 15.2 & 36.7 & 68.5
& \phantom{0}5.4 & 27.9\textsuperscript{\rlap{*}} & 71.4\textsuperscript{\rlap{*}} \\

\midrule
\multicolumn{13}{c}{\textit{Ground-Truth  (OCR + Segmentation)}} \\ \midrule

SeedX 7B PPO
& 14.5 & 27.4 & 57.8
& 35.6 & \textbf{63.1\textsuperscript{\rlap{*}}} & 81.8
& \textbf{23.5\textsuperscript{\rlap{*}}} & \textbf{49.2\textsuperscript{\rlap{*}}} & \textbf{78.9\textsuperscript{\rlap{*}}}
& 13.9 & 34.5 & 83.4 \\

Tower-Instruct 7B
& 11.0 & 31.6 & 59.2
& 28.1 & 53.2 & 75.4
& 15.1 & 34.4 & 69.2
& \textbf{23.3\textsuperscript{\rlap{*}}} & 37.5 & 83.1 \\

OmniFusion
& \textbf{18.4\textsuperscript{\rlap{*}}} & \textbf{35.0\textsuperscript{\rlap{*}}} & \textbf{62.2\textsuperscript{\rlap{*}}}
& \textbf{36.9\textsuperscript{\rlap{*}}} & 62.5 & \textbf{81.9\textsuperscript{\rlap{*}}}
& 20.4 & 38.8 & 74.0
& 16.5 & \textbf{43.5\textsuperscript{\rlap{*}}} & \textbf{84.6\textsuperscript{\rlap{*}}} \\

\bottomrule
\end{tabular}
}
\caption{Comparison of translation quality across models on the VISTRA benchmark. OCR-predicted results rely on PaddleOCR-v5. The best score within each evaluation setting is marked with *, and the best overall is \textbf{bold}.}
\label{tab:translation}
\end{table*}

\section{Experiments}
\subsection{Evaluation Data \& Metrics}
Since no dataset directly provides lecture slides with ground-truth translations, summaries, and QA pairs, we evaluate our approach on established benchmarks that approximate these tasks. For image translation, we use the VISTRA benchmark \citep{salesky2024benchmarking}, which contains real-world images such as street signs with ground-truth OCR and translations for English → {German, Chinese, Russian, Spanish}.  OCR performance is measured using Character Error Rate (CER), Term Error Rate (TER; \citealt{snover2006study}), and latency. Translation quality is evaluated with BLEU, ChrF using SacreBLEU\footnote{\scriptsize\texttt{nrefs:1|case:mixed|eff:no|tok:13a|smooth:exp|version:2.3.1}} \citep{post2018call}, and COMET\footnote{\texttt{Unbabel/wmt22-comet-da}} \citep{rei-etal-2022-comet}. For downstream tasks, we use the MCIF dataset of ACL talks \citep{papi2025mcifmultimodalcrosslingualinstructionfollowing} and report normalized BERTScore to evaluate generated summaries and answers.

\subsection{Image Translation}

We evaluate our complete image-translation pipeline along three dimensions: OCR accuracy, translation quality, and component runtime.

\paragraph{OCR Evaluation.}
Table~\ref{tab:ocr} summarizes OCR performance of several open-source systems and the vision LLM Qwen-2.5-VL (7B; \citealt{bai2025qwen2}). EasyOCR\footnote{https://github.com/JaidedAI/EasyOCR} performs the worst due to its lightweight and less robust design. PaddleOCR v4 and v5 achieve similar and much higher accuracy, while Qwen-2.5-VL matches PaddleOCR but suffers from very high latency (0.1s → 5s per image). Considering accuracy, latency, and language coverage, PaddleOCR v5 provides the best trade-off and is used for all subsequent experiments.

\paragraph{Translation Quality}
Table~\ref{tab:translation} presents translation results for both unimodal LLMs, Tower 7B \cite{Alves2024-ls} and SeedX, and the multimodal OmniFusion model. To evaluate the impact of input segmentation, we compare line-level segmentation (where each OCR line is treated independently), block-level segmentation (where lines are grouped within layout regions), and ground-truth OCR plus segmentation as an upper bound.

Overall, OmniFusion consistently outperforms unimodal translation in most languages, showing that visual context from images helps disambiguate short or visually grounded text, such as diagram labels or signs. Ground-truth OCR and segmentation yield the best performance, highlighting the importance of accurate text extraction and layout grouping. Block-level segmentation improves translation over line-level segmentation, confirming that coherent sentence-like units are critical for high-quality output. Unimodal translation performs better in Russian, indicating potentially less reliance on visual context in this direction.

\begin{figure}[!h]
    \centering
    \begin{subfigure}[b]{0.23\textwidth}
        \centering
        \includegraphics[width=\textwidth]{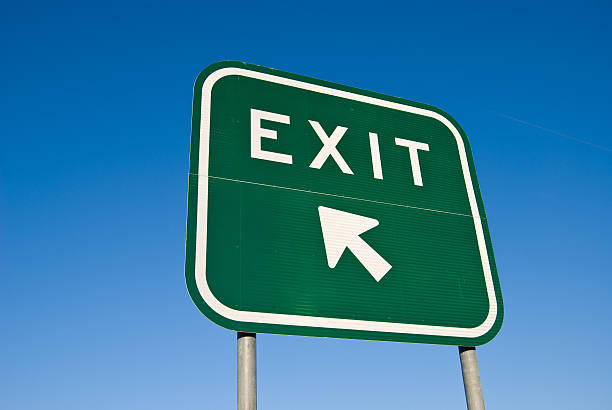}
        \caption{Original}
        \label{fig:image1}
    \end{subfigure}
    \begin{subfigure}[b]{0.23\textwidth}
        \centering
        \includegraphics[width=\textwidth]{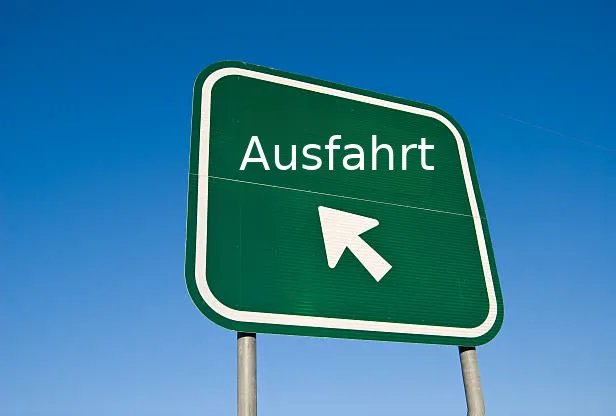}
        \caption{Translated}
        \label{fig:image2}
    \end{subfigure}
    \begin{subfigure}[b]{0.23\textwidth}
        \centering
        \includegraphics[width=\textwidth]{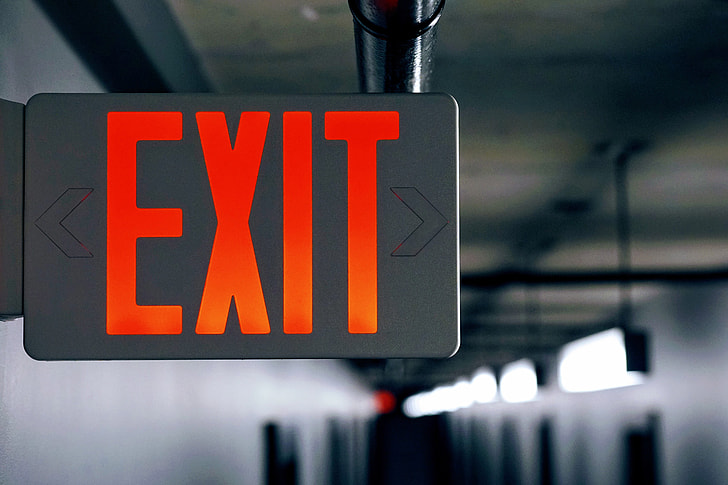}
        \caption{Original}
        \label{fig:image3}
    \end{subfigure}
    \begin{subfigure}[b]{0.23\textwidth}
        \centering
        \includegraphics[width=\textwidth]{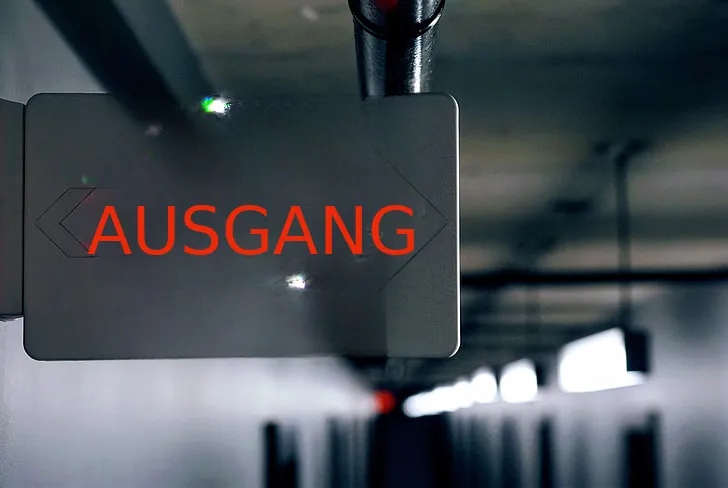}
        \caption{Translated}
        \label{fig:image4}
    \end{subfigure}
    \caption{Example illustrating that our Image Translator uses context for disambiguation. The word ``Exit'' can mean ``Ausgang'' in the context of a pedestrian exit and ``Ausfahrt'' in the context of a car exit. Our translator correctly leverages the visual context to produce different translations, even when the source text is identical in both scenarios.}
    \label{fig:exit_example}
\end{figure}

Table~\ref{tab:latency} in Appendix~\ref{sec:appendix} shows inference times for different components. Layout analysis and translation are slowest, whereas OCR and image rendering add relatively minor overhead, suggesting that optimizing efficiency for these would provide the largest latency gains. Figure~\ref{fig:exit_example} illustrates an example in which multimodal translation disambiguates text using visual context, demonstrating the practical benefit of incorporating images.


\begin{table}[t]
\centering
\resizebox{\columnwidth}{!}{
\setlength{\tabcolsep}{4pt}
\begin{tabular}{lcccc}
\toprule
Language & ST Input & LLaMA 3.1 8B & GPT OSS 20B & Mistral Small 3.2 24B \\
\midrule
\multicolumn{5}{c}{\textbf{Summarization}} \\
\midrule
\multirow{2}{*}{English} 
  & \faVolumeUp            & 18.4   & 12.1 & 18.1 \\
  & \faVolumeUp\,\faImage  & 20.5  & 12.7 & 19.7 \\
\midrule
\multirow{2}{*}{German} 
  & \faVolumeUp            & 20.6  & 18.0 & 21.7 \\
  & \faVolumeUp\,\faImage  & 23.4  & 18.9 & 24.1 \\
\midrule
\multirow{2}{*}{Italian} 
  & \faVolumeUp            & 22.5  & 18.9 & 25.4 \\
  & \faVolumeUp\,\faImage  & 24.4  & 19.7 & 26.3 \\
\midrule
\multirow{2}{*}{Chinese} 
  & \faVolumeUp            & 35.7  & 31.9 & 35.9 \\
  & \faVolumeUp\,\faImage  & 35.3  & 31.7 & 35.8 \\
\midrule
\multicolumn{5}{c}{\textbf{Question Answering}} \\
  \midrule
\multirow{2}{*}{English} 
  & \faVolumeUp            & 31.5  & 23.0  & 34.5  \\
  & \faVolumeUp\,\faImage  & 34.5  & 22.0  & 35.4  \\
\midrule
\multirow{2}{*}{German} 
  & \faVolumeUp            & 32.0  & 21.5  & 37.2  \\
  & \faVolumeUp\,\faImage  & 33.6  & 22.5  & 37.6  \\
\midrule
\multirow{2}{*}{Italian} 
  & \faVolumeUp            & 33.7  & 19.4  & 36.2  \\
  & \faVolumeUp\,\faImage  & 34.7  & 20.5  & 34.7  \\
\midrule
\multirow{2}{*}{Chinese} 
  & \faVolumeUp            & 35.8  & 30.5  & 32.4  \\
  & \faVolumeUp\,\faImage  & 35.4  & 30.0  & 32.7  \\
\midrule
\bottomrule
\end{tabular}
}
\caption{Summarization and Question Answering performance of different LLMs on the MCIF test dataset based on translations of the presentations with OmniFusion. Reported is BERTScore ($\uparrow$), rescaled with the baseline. \faVolumeUp: Audio only, \faVolumeUp\,\faImage: Audio + Image.}
\label{tab:sum-bertscore}
\end{table}

\subsection{Downstream Tasks}

We analyze how downstream performance on the MCIF benchmark \cite{papi2025mcifmultimodalcrosslingualinstructionfollowing}, specifically for Summarization and Question Answering, is affected when the transcript used as context is generated by the multimodal speech-translation system. Using the task instructions provided by MCIF, we prompt each evaluated model directly with the translated talk transcript produced by our pipeline. We evaluate three LLMs: LLaMA 3.1 8B \cite{Grattafiori2024-sv}, GPT-OSS 20B \cite{agarwal2025gpt}, and Mistral-Small 3.2 24B \cite{mistral_2023} \footnote{\url{https://mistral.ai/news/mistral-small-3-1}} \footnote{\url{https://huggingface.co/mistralai/Mistral-Small-3.2-24B-Instruct-2506}}. This setup allows us to measure how using audio-only transcripts compared to multimodal transcripts that also incorporate slide information influences downstream task performance.

 \paragraph{Summarization.} As shown in \autoref{tab:sum-bertscore}, summaries generated from audio+image input (\faVolumeUp\,\faImage) consistently outperform those based on audio-only (\faVolumeUp) across most languages and models, even though the summarization models are text-only. The gains are most pronounced in English, German and Italian, while results for Chinese slightly degraded. We presume this is because English domain terminology appears in references for Latin-alphabet languages, while the lexical distance between English and Chinese prevents the models from consistently benefiting from additional context provided in English language.

\paragraph{Question Answering.} In most settings, the results for QA are slightly better when incorporating visual context, though the gains are much less pronounced compared to summarization. In most cases, we observe small gains but also performance regression in four out of twelve language--model combinations. We assume that we do not see higher improvements and regression because the LLM does not receive the image data itself but just the (through multimodality improved) textual context which is not enough for the model to answer the questions more reliably.

\section{Related Work}

 Streaming ST has been extensively studied in the last decade \citep{machavcek2023turning, guo2025streamuni, Papi2025-mm}. Several lecture translation tools have also leveraged ST \citep{cho2013real,niehues2016dynamic,son2020low,muller2016evaluation, dessloch2018kit, huber2023end}, but these systems primarily rely on audio input. In contrast, our work extends lecture translation to multimodal input, incorporating visual cues from slides, and multimodal output, producing translated audio and slides in multiple languages.

Image-to-image translation remains relatively under-explored. Several research works focus on road sign translation \citep{gao2001text,yang2001automatic, zhang2002automatic, chen2002automatic, chen2004automatic} facing many similar challenges to translating images in academic slies. Interest in this area is growing with the availability of larger datasets \citep{zuo-etal-2025-inimagetrans, li-etal-2025-mit, zhuang2025patimt}, but most existing work focuses solely on text translation within images, without addressing the aligned re-rendering of the visual content. An initial step in this direction is \citep{tian2025prim}, which explicitly models the rendering process. Our image translation pipeline provides a modular foundation, enabling researchers to integrate models at any stage from OCR to translation and rendering, without needing to implement additional components.

\section{Conclusion}
This paper presents a multimodal, multilingual lecture translation system that leverages multiple input modalities to generate translations across different output modalities. Future work includes conducting human evaluations to assess the quality of translated slides and audio, enabling targeted improvements to the system.

\section*{Limitations}

To assess the effectiveness of our slide translation, we use the VISTRA benchmark as a proxy. However, this benchmark does not fully reflect translation quality in the lecture domain, nor does it allow us to evaluate the quality of rendered slides. Human evaluation is therefore needed to assess the rendering quality of translated slides, including layout preservation and visual coherence. For SUM and QA, we conduct evaluation only after the entire talk has been translated, which does not accurately simulate a live lecture scenario. Benchmarks with questions aligned to the lecture timeline would provide more realistic and informative evaluations for our use-case.

\section*{Acknowledgments}
The research leading to these results was supported by  European Union’s Horizon Europe programme grant agreement No. 101213369 (DVPS) and No. 101135798 (Meetween),
The German Federal Ministry of Education, Research
(BMBF) under the Robotics Institute Germany (RIG) and "How is AI Changing Science? Research in the Era of Learning Algorithms" (HiAICS) project.

\bibliography{custom}

\appendix

\section{Appendix}
\label{sec:appendix}

\begin{table}[!ht]
\centering
\begin{tabular}{@{}cc@{}}
\toprule
Step                   & Time (seconds) \\ \midrule
OCR                    & 0.46           \\
Layout Analysis        & 2.93           \\
Multimodal Translation & 3.10           \\
Inpainting             & 0.42           \\
Drawing                & 0.18           \\ \bottomrule
\end{tabular}
\caption{Inference time for each step in the pipeline for translating the image shown in Figure \ref{fig:pipeline}.}
\label{tab:latency}
\end{table}

\subsection{Text-to-Speech}
\label{sec:tts}

\begin{figure*}
    \centering
    \includegraphics[width=\textwidth]{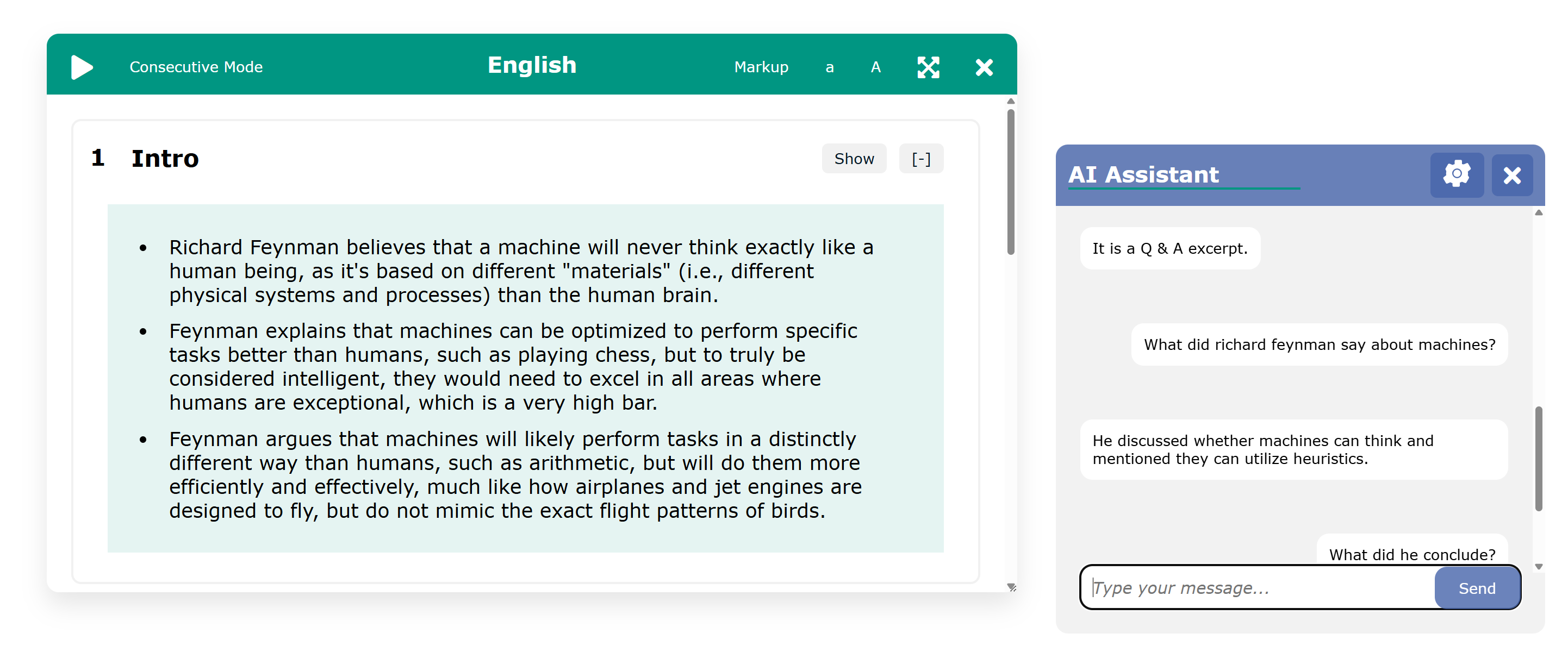}
    \caption{Summarization and Question Answering user interface. The summaries are shown for each chapter in all languages.}
    \label{fig:sumqa}
\end{figure*}

In Figure~\ref{fig:sumqa}, the interface of the TTS output can be seen. It is possible to select between the simultaneous and consecutive modes. The simultaneous mode can be used when listening to the TTS output via headphones during the talk. The consecutive mode is suitable in dialog scenarios where the TTS output is paused as long as the system recognizes speaker input. We use the VITS/VITS2 \cite{kim2021conditional, kong2023vits2} and Kokoro-82M to generate audio together with a rule-based streaming algorithm to segment input text into segments.

\subsection{User Interface Screenshots}

\begin{figure*}[t]
    \centering
    \begin{minipage}{0.98\textwidth}
        \centering
        \textbf{(a) English translation with segmentation into multiple chapters.}\\[4pt]
        \includegraphics[width=\textwidth]{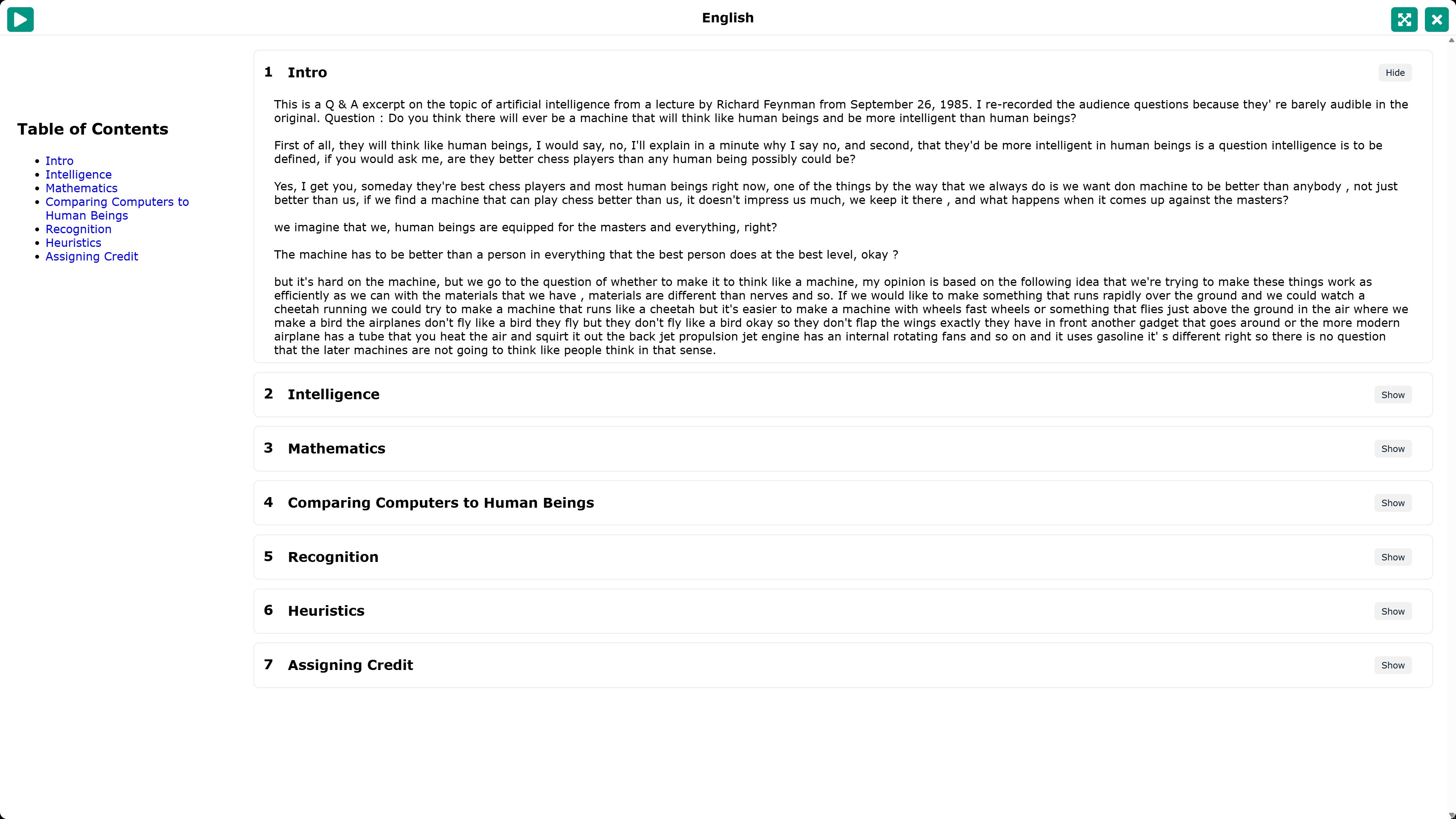}
    \end{minipage}

    \vspace{1em}

    \begin{minipage}{0.98\textwidth}
        \centering
        \textbf{(b) German translation with segmentation into multiple chapters.}\\[4pt]
        \includegraphics[width=\textwidth]{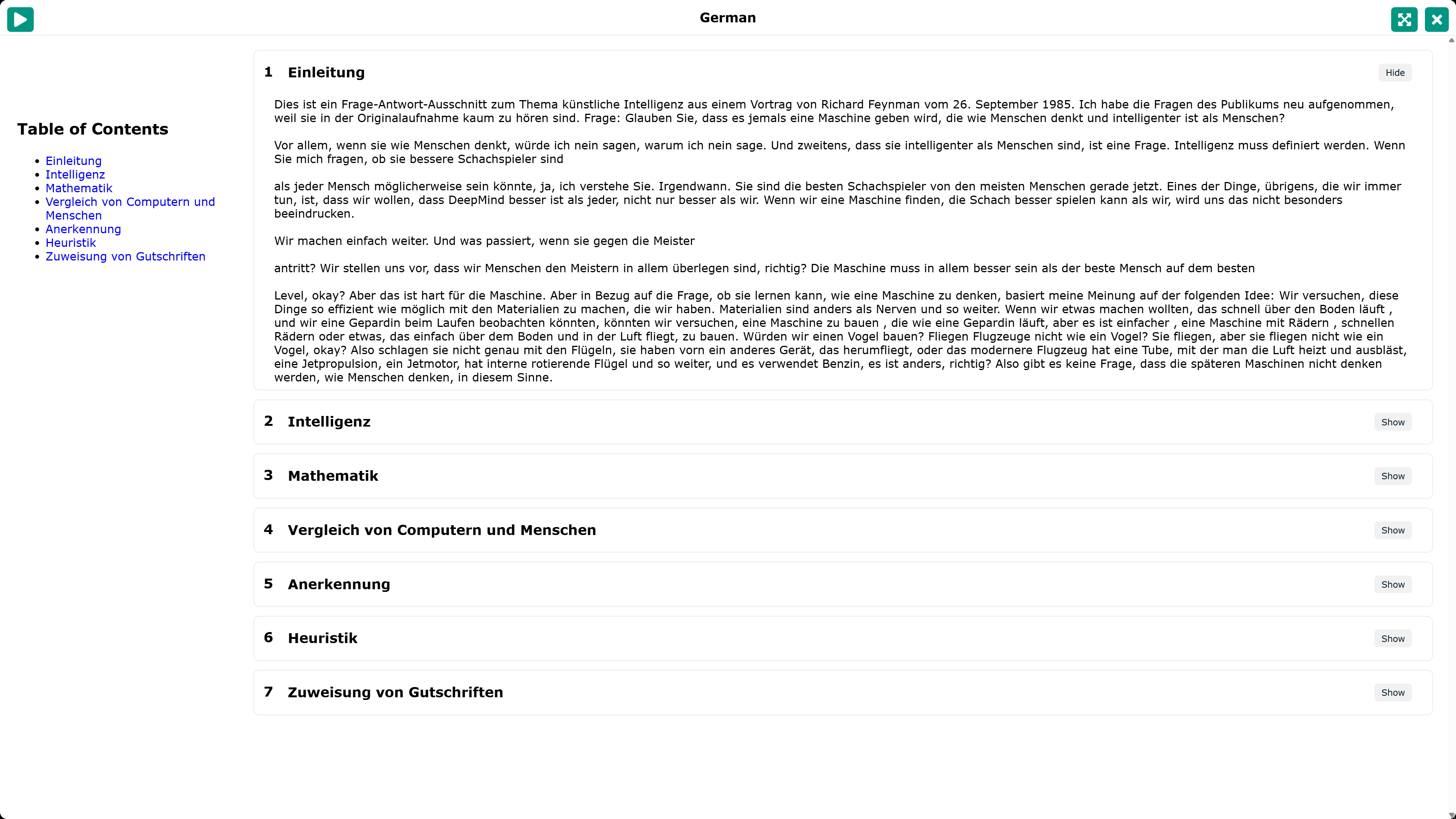}
    \end{minipage}

    \caption{Translations of the YouTube video \textbf{“Richard Feynman: Can Machines Think?”}
    (\url{https://www.youtube.com/watch?v=ipRvjS7q1DI}).  
    Subfigure (a) shows the English version; subfigure (b) shows the German version.}
    \label{fig:feynman_translations}
\end{figure*}

\begin{figure*}
    \centering
    \includegraphics[width=\textwidth]{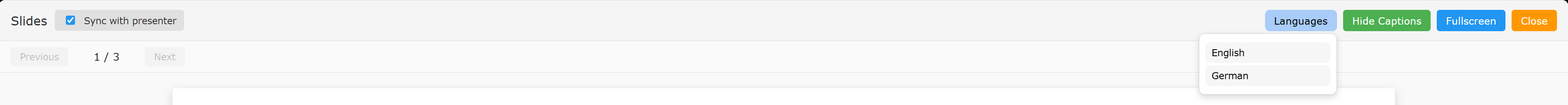}
    \caption{
    Slide viewer interface with multilingual navigation options. Users can switch between languages, browse slides independently of the presenter through an out-of-sync mode, and subsequently use the sync toggle to realign with the live presentation.
    }
    \label{fig:slideviewerui}
\end{figure*}

\begin{figure*}
    \centering
    \includegraphics[width=\textwidth]{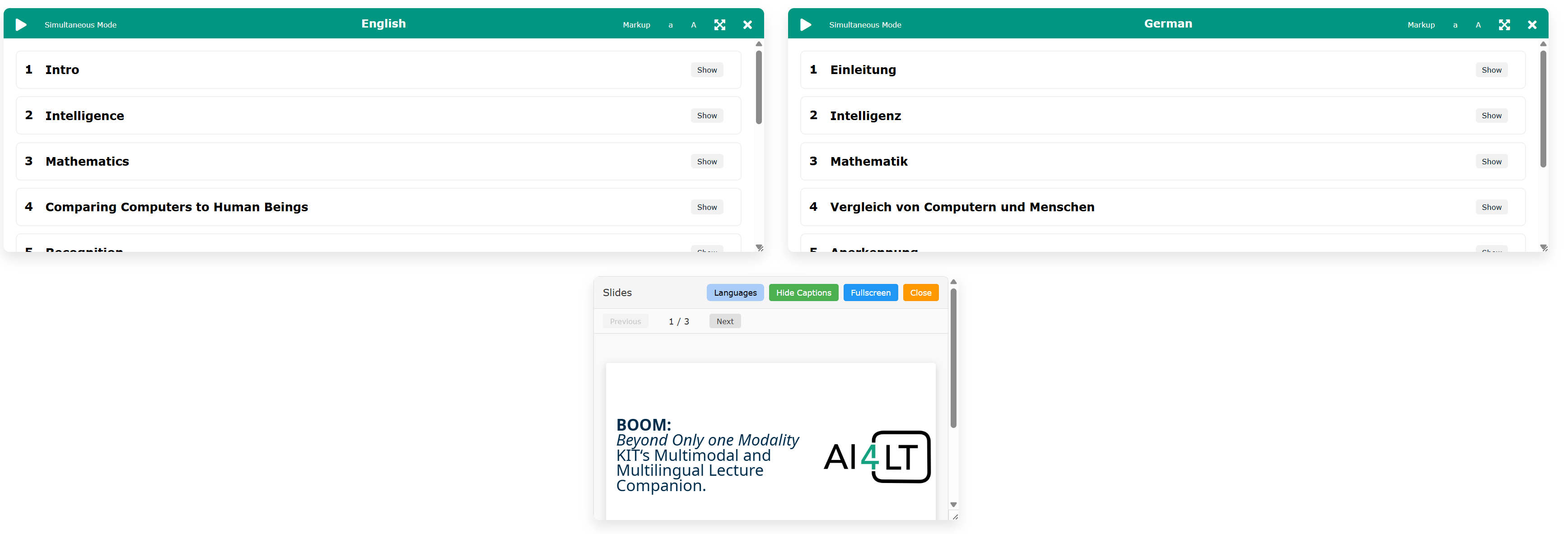}
    \caption{
    The interface also allows to see the translations in multiple languages along with the current slide.
    }
    \label{fig:window}
\end{figure*}

\begin{figure*}
    \centering
    \includegraphics[width=\textwidth]{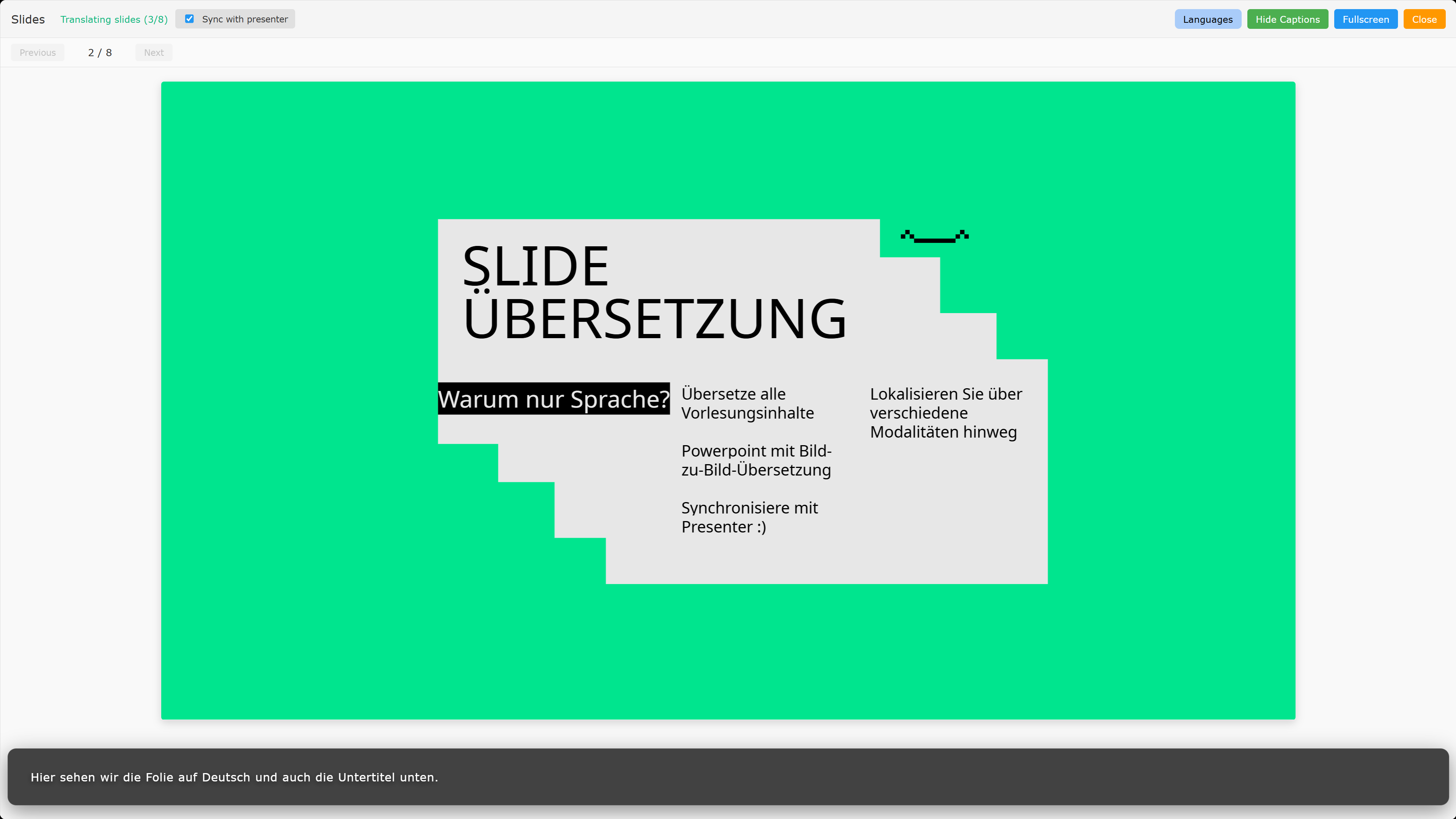}
    \caption{
    Participant full screen view of the slide interface showing slides with caption overlay in German.
    }
    \label{fig:clientview_de}
\end{figure*}

\end{document}